\documentclass[10pt,twocolumn,letterpaper]{article}

\usepackage{cvpr}              

\usepackage[dvipsnames]{xcolor}
\definecolor{cvprblue}{rgb}{0.21,0.49,0.74}
\usepackage[pagebackref,breaklinks,colorlinks,citecolor=cvprblue]{hyperref}

\usepackage{pifont}
\newcommand{\cmark}{\ding{51}}%
\newcommand{\xmark}{\ding{55}}%
\usepackage{threeparttable}
\usepackage{multirow}
\usepackage{multicol}
\usepackage{color,colortbl}

\definecolor{Gray}{gray}{0.90}
\newcolumntype{a}{>{\columncolor{Gray}}c}
\definecolor{lightgreen}{RGB}{100,220,100}
\definecolor{darkgreen}{RGB}{30,150,30}
\definecolor{darkblue}{RGB}{0,0,127}
\DeclareMathAlphabet{\mymathbb}{U}{BOONDOX-ds}{m}{n}

\def\paperID{6278} 
\def\confName{CVPR}
\def\confYear{2024}

\title{LEAD: Learning Decomposition for Source-free Universal Domain Adaptation}

\author{Sanqing Qu$^{1}$\ , Tianpei Zou$^{1}$, Lianghua He$^{1}$,  Florian Röhrbein$^{2}$,\\
Alois Knoll$^{3}$, Guang Chen$^{1}$\thanks{Corresponding author: guangchen@tongji.edu.cn}\ , Changjun Jiang$^{1}$\\
{\small $^{1}$Tongji University, $^{2}$Chemnitz University of Technology,}
{\small $^{3}$ Technical University of Munich}\\
}

\begin{document}
\maketitle
\begin{abstract}
\vspace{-0.05in}
Universal Domain Adaptation (UniDA) targets knowledge transfer in the presence of both covariate and label shifts. Recently, Source-free Universal Domain Adaptation (SF-UniDA) has emerged to achieve UniDA without access to source data, which tends to be more practical due to data protection policies. The main challenge lies in determining whether covariate-shifted samples belong to target-private unknown categories. Existing methods tackle this either through hand-crafted thresholding or by developing time-consuming iterative clustering strategies. In this paper, we propose a new idea of LEArning Decomposition (LEAD), which decouples features into source-known and -unknown components to identify target-private data. Technically, LEAD initially leverages the orthogonal decomposition analysis for feature decomposition. Then, LEAD builds instance-level decision boundaries to adaptively identify target-private data. Extensive experiments across various UniDA scenarios have demonstrated the effectiveness and superiority of LEAD. Notably, in the OPDA scenario on VisDA dataset, LEAD outperforms GLC by 3.5\% overall H-score and reduces 75\% time to derive pseudo-labeling decision boundaries. Besides, LEAD is also appealing in that it is complementary to most existing methods. The code is available at \url{https://github.com/ispc-lab/LEAD}.
\end{abstract}

\vspace{-0.15in}
\section{Introduction}
\label{sec:intro}
\par Deep neural networks (DNNs) have yielded impressive results in a variety of computer vision tasks~\cite{ren2015_faster_rcnn, he2016_resnet, dosovitskiy2020_vit, kirillov2023_segment_anything}. However, DNNs often fail to generalize well to new domains due to the discrepancy between training (source) and test (target) data distributions (i.e., covariate shift). Such a shift poses significant challenges for some safety-critical applications like autonomous driving~\cite{chen2017_da_seg, cordts2016_cityscapes}, or medical imaging~\cite{dou2018_da_medical, liu2022_medical_sfda}. Unsupervised Domain Adaptation (DA)~\cite{ganin2016_dann, kouw2019_da_review} aims to address this challenge by transferring task-specific knowledge from a labeled source domain to an unlabeled target domain. Despite promising results, most existing favorable DA methods~\cite{ganin2016_dann, hoffman2018_cycada, saito2018_mcd, liang2020_shot, qu2022_bmd} assume that the label spaces are identical across the source and target domain, thus being only applicable to closed-set scenarios and limiting applicability to practical applications.

\par To handle generalized cases, Universal Domain Adaptation (UniDA)~\cite{saito2020_unida_dance} was proposed to allow all potential label shifts between source and target domains. Different from existing methods specialized for Partial Domain Adaptation (PDA)~\cite{cao2018_pda, ETN}, Open-set Domain Adaptation (OSDA)~\cite{panareda2017open, saito2018_osbp}, or Open-Partial Domain Adaptation (OPDA)~\cite{saito2021_ovanet, you2019_uan}, in UniDA, we have no prior knowledge about the label shift, e.g., information on matched common classes or the number of categories in the target domain. Nonetheless, most existing UniDA works~\cite{saito2020_unida_dance, li2021_dcc, chen2022_gate} necessitate concurrent access to source and target data. This requirement becomes increasingly impractical in light of strict data protection policies~\cite{voigt2017_eu_gdpr}. In this paper, we focus on Source-free Universal Domain Adaptation (SF-UniDA)~\cite{liang2021_umad, qu2023_glc}, where only a pre-trained source model is provided for knowledge transfer rather than labeled source data.

\begin{figure}
    \centering
    \vspace{-0.10in}
    \includegraphics[width=0.999\linewidth]{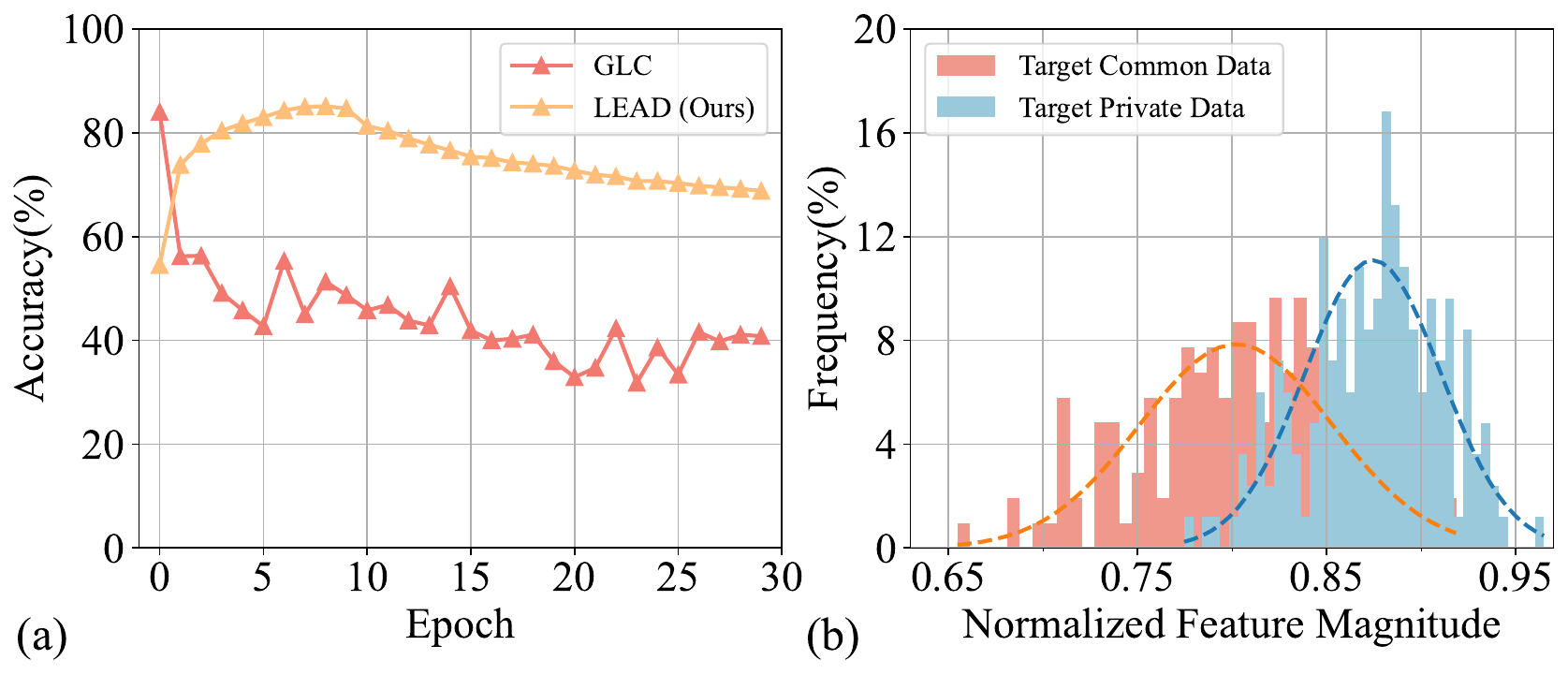}
    \vspace{-0.28in}
    \caption{(a) Pseudo-label accuracy curves for target-private data in the OPDA scenario on VisDA dataset. An observation is that clustering-based GLC does not work well in deriving decision boundaries for target-private data. This may be due to the curse of dimensionality, leading K-means to faintly discerning clusters. (b) Frequency distribution for the normalized feature magnitude of target data in the source-unknown space (orthogonal complement of the space spanned by source model weights). Features are on task ($A\rightarrow D$) in the OPDA scenario of Office-31. The results show that target-private data are expected to involve more components from source-unknown space, even with covariate shifts.
    }
    \vspace{-0.25in}
    \label{fig:intro_fig}
\end{figure}

\par The primary objective for UniDA is to recognize target data belonging to common categories and separate them from target-private ``unknown" data. Existing methods have designed various criteria based on instance-level predictions, e.g., entropy-based~\cite{you2019_uan, saito2020_unida_dance, feng2021_osht}, confidence-based~\cite{kundu2020_usfda, fu2020_cmu, saito2021_ovanet}. However, defining a universal threshold for all scenarios is challenging and often sub-optimal. To mitigate tedious parameter-tuning, some methods start to perform source and target consensus clustering to achieve common and private data separation~\cite{li2021_dcc, chen2022_gate}. Considering that SF-UniDA is inaccessible to source data, GLC~\cite{qu2023_glc} devises an iterative one-vs-all clustering strategy to realize target-private data identification. Regardless of progress, iterative clustering can be computationally intensive.  Besides, due to the curse of dimensionality~\cite{bellman1966_dynamic_programming}, methods like K-means are prone to overfitting and faintly discerning clusters. This can be implied to some degree from Figure~\ref{fig:intro_fig} (a).

\par In this paper, we propose to address the SF-UniDA problem from a new perspective, i.e., feature decomposition. Our launching point is that, even though target data might have shifted in the feature space, target-private data from unknown categories are likely to exhibit a higher proportion of components from the orthogonal complement (source-unknown) space spanned by source model weights. The frequency distribution of normalized feature magnitudes in the source-unknown space depicted in Figure~\ref{fig:intro_fig} (b) basically demonstrates our argument, which has been rarely explored in the literature. To materialize our idea, we propose the LEArning Decomposition (LEAD) framework. In particular, LEAD initially leverages the orthogonal decomposition to construct source-known and -unknown space. The projection on source-unknown space is then regarded as the discriminative representation with respect to target-private categories. Different from existing thresholding methods introducing a fixed global hand-crafted threshold for all scenarios, LEAD deliberately considers the distance to the target prototypes and source anchors to establish instance-level decision boundaries. We analyze and evaluate our LEAD under a variety of SF-UniDA scenarios, ranging from PDA, OSDA, to OPDA. Extensive experiments demonstrate the effectiveness and superiority.

\par The main contribution is the proposal of a LEArning Decomposition (LEAD) framework for source-free universal domain adaptation (SF-UniDA). This solution leads to elegant views for identifying target-private unknown data without tedious tuning thresholds or relying on iterative unstable clustering. Remarkably, in the OPDA scenario on VisDA, LEAD attains an H-score of 76.6\%, surpassing GLC~\cite{qu2023_glc} by 3.5\%. Besides, LEAD is complementary to most existing SF-UniDA methods. For instance, in the OPDA scenario on Office-Home, LEAD advances UMAD~\cite{liang2021_umad} with H-score improvement from 70.1\% to 78.0\%.

\section{Related Work}
\noindent \textbf{Unsupervised Domain Adaptation.} 
When DNNs encounter data from a distribution that differs from their training data, inevitable performance degeneration is commonly observed. To address this issue, domain adaptation (DA) ~\cite{ganin2016_dann, kouw2019_da_review} has been introduced, leveraging labeled data from source domains to train DNNs for unlabeled target domains in a transductive learning manner. Existing DA methods can be broadly categorized into two paradigms, namely moment matching~\cite{dacs_da_seg, wasserstein, mmd} and adversarial learning~\cite{ganin2016_dann, cycada, da_object_1}. Regardless of their effectiveness in various applications such as object recognition~\cite{liang2020_shot, ganin2016_dann, qu2022_bmd}, semantic segmentation~\cite{dacs_da_seg, cycada}, and object detection~\cite{da_object_1, da_object_2}, most existing approaches assume that label spaces are identical across the source and target domains, limiting their applicability.

\noindent \textbf{Universal Domain Adaptation.} There have been some approaches introduced to deal with label-shift scenarios, such as open-set domain adaptation (OSDA)~\cite{saito2018_osbp, panareda2017open, liu2022_psdc}, partial-set domain adaptation (PDA)~\cite{cao2018_pda, ETN}, and open-partial-set domain adaptation (OPDA)~\cite{you2019_uan, fu2020_cmu, saito2021_ovanet}. However, these methods are often tailored to specific scenarios and may not be readily applicable to other label shift situations. Universal domain adaptation (UniDA), on the other hand, is designed to facilitate adaptation across all potential label shift scenarios. Despite the progress in UniDA, many existing methods~\cite{saito2021_ovanet, chang2022_uniot, chen2022_gate} necessitate simultaneous access to both source and target data, making them impractical for applications with stringent data protection policies. To address this challenge, source-free universal domain adaptation (SF-UniDA)~\cite{liang2021_umad, qu2023_glc} has been proposed, wherein source data are only utilized for pre-training and are inaccessible during target adaptation. To promote the distinction between common and private data, existing methods often resort to manually defined thresholding or the introduction of global clustering techniques. Nevertheless, selecting an appropriate threshold for all scenarios can be both laborious and sub-optimal.
Additionally, K-means clustering tends to be unstable when applied to high-dimensional data due to the curse of dimensionality~\cite{bellman1966_dynamic_programming}. Our work delves into these limitations and proposes a novel and elegant solution from the perspective of feature decomposition.

\noindent \textbf{Feature Decomposition.} 
As a fundamental technique in machine learning, feature decomposition is applied to deconstruct complex data into simpler and more interpretable components. In the field of domain adaptation and domain generalization, feature decomposition has been leveraged in various studies~\cite{li2018_da_decompose, liu2022_da_decompose, chattopadhyay2020_learning, piratla2020_efficient, qu2023_mad, liu2023_d2ifln} to enhance feature alignment or facilitate domain-invariant feature learning by breaking down features into content and style components. Nonetheless, the content and style decomposition design is less applicable for UniDA, as the primary challenge lies in distinguishing common and private data. In this paper, we approach the problem from the perspective of orthogonal feature decomposition and employ Independent Component Analysis to decompose features into two orthogonal parts, thereby facilitating the intended objective.

\begin{figure*}[ht]
    \centering
    \vspace{-0.00in}
    \includegraphics[width=0.95\textwidth]{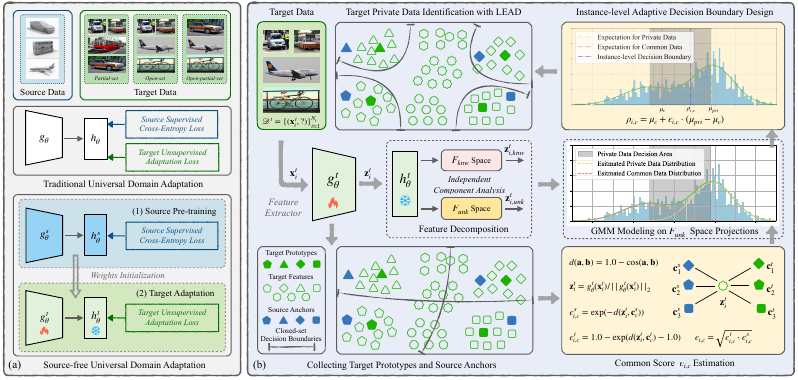}
    \vspace{-0.10in}
    \caption{
    (a) Illustrations of traditional universal domain adaptation (UniDA) and source-free universal domain adaptation (SF-UniDA). Traditional UniDA methods necessitate data from both source and target domains concurrently. In SF-UniDA, source data are solely utilized for pre-training. Adaptation is performed by harnessing target data and the source model $f^s_\theta = h^s_\theta \circ g^s_\theta$.
    (b) An overview of our Learning Decomposition (LEAD) framework. Pseudo-labeling is an important technique for UniDA and SF-UniDA. The primary objective is to recognize target data associated with common label sets and exclude data within the target-private label space. 
    Different from existing methods that perform private data identification by hand-crafted thresholding on predictions or iterative global clustering, we tackle this from the viewpoint of feature decomposition. The rationale is that despite potential shifts in the feature space, target-private data are expected to encompass more components from the orthogonal complement (source-unknown) space of the source model. Technically, LEAD first performs 
    orthogonal decomposition to decompose target features into source-known and -unknown parts, i.e., $\mathbf{z}^t_{i, knw}$ and $\mathbf{z}^t_{i, unk}$. $\lVert\mathbf{z}^t_{i, unk}\rVert_2$ is considered as an indicator for private data. Next, LEAD employs a two-component Gaussian Mixture Model to estimate the distribution of $\lVert\mathbf{z}^t_{i, unk}\rVert_2$. Thereafter, LEAD devises a metric named ``common score" $\epsilon_{i, c}$ that accounts for distances to both target prototypes and source anchors (derived from $h^t_{\theta}$) to facilitate deriving instance-level decision boundary $\rho_{i, c}$. LEAD provides an elegant solution to distinguish target-private data, mitigating the need for tedious hand-crafted threshold tuning or dependence on time-consuming iterative clustering. LEAD could also serve as a complementary approach to most existing SF-UniDA methods.
    }
    \vspace{-0.15in}
    \label{fig:framework}
\end{figure*}

\section{Methodology}
\subsection{Preliminary}
\par In this paper, we consider a generalized and challenging case in domain adaptation: universal domain adaptation (UniDA), which aims to achieve knowledge transfer in the presence of both covariate and label shifts.
In UniDA, there is a labeled source domain $\mathcal{D}^s  = \{(\mathbf{x}^s_i, \mathbf{y}^s_i)\}_{i=1}^{N_s}$ where $\mathbf{x}^s_i \in \mathcal{X}^s \subset \mathbb{R}^{X}, \mathbf{y}^s_i \in \mathcal{Y}^s \subset \mathbb{R}^C$, and an unlabeled target domain $\mathcal{D}^t = \{(\mathbf{x}^t_i, ?)\}^{N_t}_{i=1}$ where $\mathbf{x}^t_i \in \mathcal{X}^t \subset \mathbb{R}^{X}$. We refer to $\mathcal{Y}^t$ as target label sets. $\mathcal{Y} = \mathcal{Y}^s \cap \mathcal{Y}^t$ denotes the common label sets shared by source and target domains. $\bar{\mathcal{Y}}^s = \mathcal{Y}^s \setminus \mathcal{Y}$ and $\bar{\mathcal{Y}}^t = \mathcal{Y}^t \setminus \mathcal{Y}$ symbolizes the label sets private to source and target domain, individually. UniDA supposes that we have no prior knowledge about $\mathcal{Y}^t$. $\mathcal{Y}$ and $\bar{\mathcal{Y}}^t$ are also unavailable. The objective is to recognize common data belonging to $\mathcal{Y}$ and reject private data in $\bar{\mathcal{Y}}^t$. 

\par Different from traditional UniDA methods that need concurrent access to source and target data, under the source-free universal domain adaptation (SF-UniDA) setting, $\mathcal{D}_s$ is only applicable for source model pre-training and not accessible during adaptation. Assuming that the source model is represented as $f^s_{\theta} = h^s_{\theta} \circ g^s_{\theta}$, where $g^s_{\theta}: \mathbb{R}^X \rightarrow \mathbb{R}^D$ denotes the feature extractor and $h^s_{\theta}: \mathbb{R}^D \rightarrow \mathbb{R}^C$ is the classifier module. Following existing methods~\cite{liang2020_shot, qu2022_bmd},  we freeze the classifier module $h^t_{\theta} = h^s_{\theta}$ and only learn a target-specific feature extractor $g^t_{\theta}$ to realize model adaptation. Figure~\ref{fig:framework} (a) presents an illustration for UniDA and SF-UniDA.
\par The main challenge is how to distinguish common data from private data. Prior methods~\cite{saito2020_unida_dance, liang2021_umad, li2021_dcc, qu2023_glc} have designed various algorithms to encourage the separation. However, these methods perform data identification either based on hand-crafted thresholding criteria or by introducing time-consuming unstable clustering strategies. We propose to mitigate these limitations from the perspective of feature decomposition. An observation is that even though target data may have shifted in the feature space, the features of private data still include more components from the orthogonal complement (source-unknown) space of the pre-trained model. Technically, we present a novel learning decomposition (LEAD) framework. LEAD capitalizes on the orthogonal decomposition to build two orthogonal feature spaces, i.e., source-known and -unknown space. Feature projection on source-unknown space is extracted as the descriptor for private data. Then, LEAD considers the distances to both the target prototypes and source anchors to establish instance-level decision boundaries. Figure~\ref{fig:framework} (b) manifests the architecture of our LEAD.

\subsection{Orthogonal Feature Decomposition}
\par Pseudo-labeling serves as an important technique in unsupervised learning, yet many existing strategies only account for closed-set scenarios, disregarding the label-shift situations, which impairs their ability to differentiate target-private data. To fulfill the objective of common and private data separation, we propose to orthogonally decompose the feature representation into two uncorrelated parts, i.e., features associated with source-known space  $F_{knw}$ and those related to source-unknown space $F_{unk}$. This design stems from a simple inductive bias in favor of private data identification. \textit{Due to the lack of exposure to target data during source pre-training, particularly target-private data, the source-known space does not include any significant attributes of target-private data.} Subsequently, after normalization, target-private data are expected to contain more components from source-unknown space, even with covariate shifts. This hypothesis is basically demonstrated by empirical evidence, as illustrated in Figure~\ref{fig:intro_fig} (b).

\par Formally, denoting the weight of classifier $h^t_{\theta}$ is $W_{cls} \in \mathbb{R}^{C\times D}$. Source-known space $F_{knw}$ can be represented as the vector space of $\mathrm{span}\{\mathbf{w}_n\}$, where $\{\mathbf{w}_n \in \mathbb{R}^{D} |_{n=1,\dots ,C}\}$ is a group weight vectors of $W_{cls}$. Source-unknown space $F_{unk}$ is the orthogonal complement space of $F_{knw}$, i.e., $\mathbb{R}^D = F_{knw} \oplus F_{unk}$. Nevertheless, unit orthogonality in $\{\mathbf{w}_n \in \mathbb{R}^{D} |_{n=1,\dots ,C}\}$ is not guaranteed (i.e., when $n\ne m$  ($\mathbf{w}_n, \mathbf{w}_m) \ne 0$), making it hard to derive $F_{unk}$. Considering that $F_{knw}$ and $F_{unk}$ are the row space and null space of $W_{cls}$. Therefore, we employ the singular value decomposition (SVD) on $W_{cls}$ to obtain $F_{knw}$ and $F_{unk}$. Specifically, 
\begin{equation}
    \vspace{-0.02in}
    \begin{aligned}
    W_{cls} &= U\Sigma V^T, \\
    F_{knw} &= \mathrm{span} \{\mathbf{v}_n|_{n=1,\dots ,C}\},\\
    F_{unk} &= \mathrm{span} \{\mathbf{v}_n|_{n=C+1,\dots ,D}\} 
    \end{aligned}
    \vspace{-0.02in}
\end{equation}
where $\Sigma\in \mathbb{R}^{C\times D}$ is a diagonal matrix, $U\in\mathbb{R}^{C\times C}$ and $V\in\mathbb{R}^{D\times D}$ are both orthogonal unitary matrices. $\{\mathbf{v}_n | \mathbf{v}_n \in \mathbb{R}^{D}\}^{D}_{n=1}$ are columns of $V$.
\par Assuming that $\mathbf{z}_i^t$ is the feature of target data $\mathbf{x}_i^t$ extracted from feature extractor $g^t_{\theta}$ after normalization, i.e., $\mathbf{z}^t_i = g^t_{\theta}(\mathbf{x}^t_i) / ||g^t_{\theta}(\mathbf{x}^t_i)||_2$, then $\mathbf{z}^t_i$ can be represented as the weighted summation of two orthogonal bases:
\vspace{-0.05in}
\begin{equation}
    \begin{aligned}
    \mathbf{z}^t_i = \underbrace{\sum_{n=1}^{C} l_n \cdot \mathbf{v}_n}_{\mathbf{z}^t_{i, knw}} + \underbrace{\sum_{n=C+1}^{D} l_n \cdot \mathbf{v}_n}_{\mathbf{z}^t_{i, unk}}
    \end{aligned}
    \vspace{-0.10in}
\end{equation}
where ${||\mathbf{z}^t_{i, knw}||_2}^{2} + {\lVert\mathbf{z}^t_{i, unk}\rVert_2}^{2} = 1$, $\mathbf{z}^t_{i, knw}$ and $\mathbf{z}^t_{i, unk}$ are the projections of $\mathbf{z}^t_i$ on $F_{knw}$ and $F_{unk}$, respectively. $l_n \in \mathbb{R}$ denotes the projection weight of $\mathbf{z}^t_i$ onto the unit base $\mathbf{v}_n$.
This procedure is commonly recognized as orthogonal decomposition and has been extensively studied in the context of Independent Component Analysis (ICA).

\subsection{Adaptive Separation for Private Data}
\par Intuitively, enabled by feature decomposition, we could differentiate between common and private data by examining the dominance of $\mathbf{z}^t_{i, unk}$ over $\mathbf{z}_i$. However, such a straightforward approach is not applicable to SF-UniDA due to the impact of distribution covariate shift, which often deteriorates the significance of $\mathbf{z}^t_{i, knw}$ for common data, leading to the prevalence of $\mathbf{z}^t_{i, unk}$ in the representation. Additionally, the feature extractor inevitably captures representations that are independent of $F_{knw}$, including background contexts~\cite{beery2018_recognition}, object textures~\cite{geirhos2018_texture_bias}, and high-frequency patterns~\cite{wang2020_high_frequency}, further exacerbating this issue.

\par To solve this, one feasible solution is to estimate the distribution of $\lVert\mathbf{z}^t_{i, unk}\rVert_2$ to effectively calibrate these disturbances. Our observations have revealed that the empirical distributions of $\lVert\mathbf{z}^t_{i, unk}\rVert_2$ exhibit a bimodal pattern, with the presence of two distinct peaks typically indicating the modes of common and private data. Consequently, we employ a two-component Gaussian Mixture Model (GMM) to estimate the distribution $\lVert\mathbf{z}^t_{i, unk}\rVert_2$, with the components featuring a lower mean representing the common data and those with a higher mean corresponding to the private data. In particular, we denote $\mu_{com}$ and $\mu_{pri}$ as the expectation for common and private data, where $\mu_{com} < \mu_{pri}$. 

\par In the light of $\mu_{com}$ and $\mu_{pri}$, an intuitive practice is to treat all common and private categories equally. Accordingly, common and private data can be separated by establishing $(\mu_{com} + \mu_{pri}) / 2$ as the threshold,  where data with $\lVert\mathbf{z}^t_{i, unk}\rVert_2$ below the threshold are recognized as common data, while those above are identified as private data. Despite its reasonable effectiveness, it discounts the inconsistency of covariate shifts across each category. Moreover, it fails to account for the variability in covariate shifts even among samples from the same category. These considerations motivate us to devise an instance-level decision strategy to achieve accurate and effective data identification.

\par Concretely, we first leverage top-$K$ sampling to construct target prototypes, denoted as $\{\mathbf{c}^t_c \in \mathbb{R}^D |_{c=1,\dots,C}\}$ for each common category, adhering to the procedure in~\cite{qu2023_glc, liu2023_deja}. Similar to~\cite{qu2023_glc}, we empirically set $K = N_t / \tilde{C}_t$, where $N_t$ represents the amount of target data, and $\tilde{C}_t$ signifies the estimated count of target categories. Diverging from~\cite{qu2023_glc}, we additionally collect source anchors from $W_{cls}$, denoted as $\{\mathbf{c}^s_c \in \mathbb{R}^D |_{c=1,\dots,C}\}$ to assist us handle scenarios involving source-private categories. The construct of target prototypes through top-$K$ sampling fails to consider these scenarios, making it prone to misidentify private data as common. Due to limited space, further details regarding the target prototype construction are included in the appendix.

\par Thereafter, we propose a novel metric, {common score} $\epsilon_{i, c}$, which deliberately considers the distance of $\mathbf{z}^t_i$ to the target prototype $\mathbf{c}_c^t$ and source anchor $\mathbf{c}_c^s$ to facilitate us in deriving instance-level decision boundaries. Formally, this metric is defined as:
\vspace{-0.05in}
\begin{equation}
    \begin{aligned}
        d(\mathbf{a}, \mathbf{b}) &= 1.0 - \mathrm{cos}(\mathbf{a}, \mathbf{b}),\\
        \epsilon_{i, c}^t &= 1.0 - \mathrm{exp}(d(\mathbf{z}^t_i, \mathbf{c}_c^t) - 1.0),\\
        \epsilon_{i, c}^s &= \mathrm{exp}(-d(\mathbf{z}^t_i, \mathbf{c}_c^s)),\\
        \epsilon_{i, c} &= \sqrt{\epsilon_{i, c}^t \cdot \epsilon_{i, c}^s}
    \end{aligned}
    \vspace{-0.05in}
\end{equation}
where $\epsilon_{i, c}^t$ and $\epsilon_{i, c}^s$ denote the common score from target prototype and source anchor viewpoint, respectively. Both $\epsilon_{i, c}^t$ and $\epsilon_{i, c}^s$ are clipped to the [0, 1] before fusion. We then utilize geometric mean to calculate fused common score, due to its advantage in multi-criteria decision-making. Note that we leverage different formulas to calculate $\epsilon_{i, c}^t$ and $\epsilon_{i, c}^s$. Under the same condition, i.e., $d(\mathbf{z}^t_i, \mathbf{c}_c^s) = d(\mathbf{z}^t_i, \mathbf{c}_c^t)$, $\epsilon_{i, c}^s$ is greater than $\epsilon_{i, c}^t$. The spirit behind is that instances exhibiting proximity to source anchors are likely to be common data. Conversely, instances in close proximity to target prototypes do not guarantee their recognition.

\par Given the computed common score $\epsilon_{i, c}$ and the estimated distribution for $\lVert\mathbf{z}^t_{i, unk}\rVert_2$, we introduce the following pseudo-labeling strategy to effectively distinguish between common and private data instances:
\vspace{-0.05in}
\begin{align}
    \rho_{i, c} &= \mu_{c} + \epsilon_{i, c}\cdot (\mu_{pri} - \mu_c), \nonumber \\
    \tilde{\mathbf{y}}^t_i &= \left\{
    \begin{aligned}
     &\text{unknown}, &\text{if $\lVert\mathbf{z}^t_{i, unk}\rVert_2 \ge \rho_{i, c}$}\\
    &\mymathbb{1}\left(\text{argmax}(\epsilon_{i})\right), &\text{if $\lVert\mathbf{z}^t_{i, unk}\rVert_2 < \rho_{i, c}$}
    \end{aligned}
    \right.
\label{equ:psd_label}
\vspace{-0.05in}
\end{align}
where $\mu_{c}$ denotes the expectation of $\lVert\mathbf{z}^t_{i, unk}\rVert_2$ for the $c$-th category, derived using the identical top-$K$ sampling technique employed during the construction of the target prototypes.
$\rho_{i, c}$ represents the instance-level decision boundary, which accounts for the potential covariate shifts across each category, as well as the data instances corresponding to both target prototypes and source anchors. $\mymathbb{1}$ denotes the one-hot encoding operator, $\tilde{\mathbf{y}}^t_i$ is the pseudo-label for $\mathbf{x}^t_i$.

\subsection{Optimization Objectives}
\par To encourage the separation between common and private data, the optimization process involves three distinct objectives, including pseudo-label learning, feature decomposition regularization, and feature consensus regularization.
\\
\noindent \textbf{Pseudo-label Learning.} With the pseudo labels derived from Eq.~\ref{equ:psd_label}, we apply the cross-entropy loss for model adaptation. Nevertheless, instead of assigning equal importance to all pseudo-labels, we introduce a Student's $t$ distribution to characterize the certainty associated with each pseudo-label, taking into account the distance between $\lVert\mathbf{z}^t_{i, unk}\rVert_2$ and $\rho_{i, c}$. Specifically,
\vspace{-0.05in}
\begin{align}
    &\tau_{i}^t \propto 1 - \left(1 + \frac{(\rho_{i, \kappa} - \lVert \mathbf{z}^t_{i, unk}\rVert_2)^2}{\alpha}\right)^{-\frac{\alpha +1}{2}}, \nonumber\\
    &\mathcal{L}_{ce} = - \frac{1}{N}\sum^{N}_{i=1}\tau_{i}^t \cdot\sum^{C}_{c=1} \tilde{y}_{i,c}^t \log \delta_c(f^t_\theta(\mathbf{x}^t_i))
\end{align}
where  $\tau_i^t$ denotes the certainty, $\kappa = \text{argmax}(\epsilon_{i, c})$, $\alpha = $ 1e-4 by default. $\delta_c(f^t_\theta(\mathbf{x}_i^t))$ is the $c$-th soft-max probability for $\mathbf{x}_i^t$. $\tilde{y}_{i,c}^t$ corresponds to $c$-th one-hot encoded pseudo-label for $\tilde{\mathbf{y}}^t_i$. For those instances pseudo-labeled as private data, we refrain from introducing an additional $C+1$-th class, instead employing a uniform distribution to represent them. 
\\
\noindent \textbf{Feature Decomposition Regularization.} To promote the discriminability of $\mathbf{z}^t_{i, unk}$ for identifying private data, we impose a feature decomposition regularizer. Concretely,
\vspace{-0.05in}
\begin{align}
    &p_i = \frac{\exp(\lVert \mathbf{z}^t_{i, unk}\rVert_2)}{\exp(\lVert \mathbf{z}^t_{i, unk}\rVert_2) + \exp(\lVert \mathbf{z}^t_{i, knw}\rVert_2)},\\
    &\mathcal{L}_{reg} = - \frac{1}{N}\sum_{i=1}^N\tau_i^t\cdot \left(\hat{y}_i^t\log(p_i) + (1 - \hat{y}_i^t)\log (1 - p_i)\right)  \nonumber
\end{align}
$\hat{y}_{i}^t = 1 \mathrm{or}\ 0$ means pseudo-labeled as private/common data.
\\
\noindent \textbf{Feature Consensus Regularization.} 
Recent studies~\cite{yang2021_gsfda, yang2021_nrc, qu2023_glc} in SFDA and SF-UniDA observe that incorporating consensus regularization with nearest neighbors in feature space could assist models in achieving more stable performance. Following them, we integrate the objective $\mathcal{L}_{con}$ into LEAD. More details are included in the supplementary.
\\
\noindent \textbf{Overall Optimization Objective.} Taking into account the discussions above, the eventual loss function is defined as:
\vspace{-0.05in}
\begin{equation}
    \mathcal{L} = \lambda\cdot \mathcal{L}_{ce} + \mathcal{L}_{reg} + \mathcal{L}_{con}
\end{equation}
where $\lambda > 0$ is a trade-off hyper-parameter.\\
\vspace{-0.05in}
\begin{table}[t]
\centering
\caption{Details of class split. $\mathcal{Y}$, $\bar{\mathcal{Y}}_s$, and $\bar{\mathcal{Y}}_t$ signifies the common class, source-private class, and target-private class, respectively.}
\vspace{-0.15in}
\addtolength{\tabcolsep}{4.5pt}
\resizebox{0.45\textwidth}{!}{
\begin{tabular}{l|ccc}
\toprule
\multirow{2}{*}{Dataset} & \multicolumn{3}{c}{Class Split($\mathcal{Y}/ \bar{\mathcal{Y}}_s/ \bar{\mathcal{Y}}_t$)}  \\
\cmidrule{2-4} & OPDA  & OSDA  & PDA \\ 
\midrule
Office-31~\cite{office31} & 10/10/11 & 10/0/11 & 10/21/0 \\
Office-Home~\cite{officehome} & 10/5/50 & 25/0/40 & 25/40/0 \\
VisDA~\cite{visda} & 6/3/3 & 6/0/6  & 6/6/0  \\
DomainNet~\cite{domainnet} & 150/50/145   & -       & - \\ 
\bottomrule
\end{tabular}
}
\label{tab:label_split}
\vspace{-0.25in}
\end{table}

\begin{table*}[htbp]
  \centering
  \caption{H-score (\%) comparison in OPDA scenario on Office-Home. U denotes methods applicable for all potential label-shift scenarios (i.e., OPDA, OSDA, and PDA). SF represents source data-free. CF indicates model adaptation without time-consuming K-means clustering.
  }
  \vspace{-0.10in}
  \addtolength{\tabcolsep}{0.0pt}
  \resizebox{0.99\textwidth}{!}{
    \begin{tabular}{lccc|ccccccccccccl}
    \toprule
     {Methods} & {U} & {SF}  & {CF} & Ar2Cl & Ar2Pr & Ar2Re & Cl2Ar & Cl2Pr & Cl2Re & Pr2Ar & Pr2Cl & Pr2Re & Re2Ar & Re2Cl & Re2Pr & \textbf{Avg} \\
    \hline
    CMU~\cite{fu2020_cmu} & \xmark  & \xmark & \cmark &  56.0  & 56.9  & 59.2  & 67.0  & 64.3  & 67.8  & 54.7  & 51.1  & 66.4  & 68.2  & 57.9  & 69.7  & 61.6  \\
    DANCE~\cite{saito2020_unida_dance} & \cmark & \xmark & \cmark & 61.0  & 60.4  & 64.9  & 65.7  & 58.8  & 61.8  & 73.1  & 61.2  & 66.6  & 67.7  & 62.4  & 63.7  & 63.9  \\
    DCC~\cite{li2021_dcc} & \cmark & \xmark  &\xmark & 58.0  & 54.1  & 58.0  & 74.6  & 70.6  & 77.5  & 64.3  & {73.6}  & 74.9  & {81.0} & {75.1} & 80.4  & 70.2  \\
    OVANet~\cite{saito2021_ovanet} & \xmark & \xmark &\cmark & 62.8  & 75.6  & 78.6  & 70.7  & 68.8  & 75.0  & 71.3  & 58.6  & 80.5  & 76.1  & 64.1  & 78.9  & 71.8  \\
    GATE~\cite{chen2022_gate}  & \cmark & \xmark &\cmark & 63.8  & {75.9}  & {81.4}  & {74.0}  & {72.1}  & 79.8  & 74.7  & {70.3}  & {82.7} & 79.1  & 71.5  & {81.7}  & 75.6  \\
    UniOT~\cite{chang2022_uniot} & \xmark &\xmark &\cmark & 67.3 & 80.5 & 86.0 & 73.5 & 77.3 & 84.3 & 75.5 & 63.3 & 86.0 & 77.8 & 65.4 & 81.9 & {76.6} \\
    \midrule
    \midrule
    Source-only & - &- &- & 47.3  & 71.6  & 81.9  & 51.5  & 57.2  & 69.4  & 56.0  & 40.3  & 76.6  & 61.4  & 44.2  & 73.5  & 60.9  \\
    SHOT-O~\cite{liang2020_shot} & \xmark & \cmark & \cmark & 32.9  & 29.5  & 39.6  & 56.8  & 30.1  & 41.1  & 54.9  & 35.4  & 42.3  & 58.5  & 33.5  & 33.3  & 40.7  \\
    \rowcolor{Gray} LEAD & \cmark &\cmark & \cmark  & 62.7 & 78.1 & 86.4 & 70.6 & 76.3 & 83.4 & 75.3 & 60.6 & 86.2 & 75.4 & 60.7 & 83.7 & \textbf{75.0}\\
    \hline
    UMAD~\cite{liang2021_umad} & \xmark & \cmark  & \cmark & 61.1  & 76.3  & 82.7  & 70.7  & 67.7  & 75.7  & 64.4  & {55.7}  & 76.3  & 73.2  & {60.4}  & 77.2  & 70.1  \\
    \rowcolor{Gray} + LEAD & \cmark &\cmark &\cmark & 65.2 & 82.6 & 87.4 & 78.0 & 76.5 & 87.5 & 79.8 & 65.0 & 88.4 & 79.4 & 63.0 & 83.5 & \textbf{78.0} {\textcolor{darkgreen}{\footnotesize (+7.9)}} \\
    \hline
    GLC~\cite{qu2023_glc} & \cmark  & \cmark  &\xmark & 64.3     & 78.2     & 89.8     & 63.1      & 81.7  & {89.1}     &{77.6}      & 54.2      & 88.9      & 80.7    & 54.2      & 85.9     & {75.6} \\
    \rowcolor{Gray} + LEAD & \cmark &\cmark &\xmark &65.7 &79.7 &88.7 &65.0 &80.3 &88.5 &77.6 &57.4 &88.9 &78.9 &57.8 &85.5 & \textbf{76.2} {\textcolor{darkgreen}{\footnotesize (+0.6)}}\\
    \bottomrule
    \end{tabular}
    }
  \label{tab:opda_officehome}%
  \vspace{-0.12in}
\end{table*}%

\begin{table*}[htbp]
  \centering
  \caption{H-score (\%) comparison in OPDA scenario on Office-31, VisDA, and DomainNet. Some results are cited from UMAD~\cite{liang2021_umad}.}
  \vspace{-0.10in}
  \addtolength{\tabcolsep}{-1.0pt}
  \resizebox{0.99\textwidth}{!}{
    \begin{tabular}{lccc|ccccccl|ccccccl|l}
    \toprule
    \multirow{2}[1]{*}{Methods} & \multirow{2}[1]{*}{U} & \multirow{2}[1]{*}{SF} & \multirow{2}[1]{*}{CF} & \multicolumn{7}{c|}{Office-31} & \multicolumn{7}{c}{DomainNet} & \multicolumn{1}{|l}{VisDA}\\
\cline{5-19}    &   &    &  & A2D   & A2W   & D2A   & D2W   & W2A   & W2D   & \textbf{Avg}  & {P2R} & {P2S} & {R2P} & {R2S} & {S2P} & {S2R} & \textbf{Avg} &  \textbf{S2R}\\
    \hline
    CMU~\cite{fu2020_cmu} & \xmark  & \xmark & \cmark & 68.1  & 67.3  & 71.4  & 79.3  & 72.2  & 80.4  & 73.1    & {50.8} & {{45.1}} & {52.2} & {45.6} & {44.8} & {51.0} & {48.3} & 32.9\\
    DANCE~\cite{saito2020_unida_dance} & \cmark & \xmark & \cmark & 78.6  & 71.5  & 79.9  & 91.4  & 72.2  & 87.9  & 80.3    & {21.0}  & {37.0} & {47.3} & {46.7} & {27.7}  & {21.0} & {33.5} & 42.8\\
    DCC~\cite{li2021_dcc} & \cmark  & \xmark & \xmark & {88.5}  & 78.5  & 70.2  & 79.3  & 75.9  & 88.6  & 80.2    & {56.9} & {43.7} & {50.3} & {43.3} & {44.9} & {56.2} & {49.2} & 43.0\\
    OVANet~\cite{saito2021_ovanet} & \xmark & \xmark & \cmark & 85.8  & 79.4  & 80.1  & {95.4}  & 84.0  &{94.3}  & 86.5    & {56.0} & {47.1} & {51.7} & {44.9} & {47.4}  & {57.2} & {50.7} & 53.1\\
    GATE~\cite{chen2022_gate} & \cmark & \xmark & \cmark & {87.7}  & {81.6} & 84.2  &94.8  & 83.4  & 94.1  & 87.6    & 57.4 & 48.7 & 52.8  & 47.6 & 49.5 & {56.3} & 52.1 & 56.4\\
    UniOT~\cite{chang2022_uniot} & \xmark &\xmark &\cmark & 87.0 & 88.5 & 88.4 & 98.8 & 87.6 & 96.6 & {91.1} & 59.3 & 51.8 & 47.8 & 48.3 & 46.8 & 58.3 & 52.0 & 57.3\\
    \midrule
    \midrule
    Source-only &- & - & -& 70.9  & 63.2  & 39.6  & 77.3  & 52.2  & 86.4  & 64.9    & 57.3      & 38.2      & 47.8      & 38.4      & 32.2      & 48.2      & 43.7 & 25.7\\
    SHOT-O~\cite{liang2020_shot} & \xmark & \cmark & \cmark & 73.5  & 67.2  & 59.3  & 88.3  & 77.1  & 84.4  & 75.0    & {35.0} & {30.8} & {37.2} & {28.3} & {31.9} & {32.2} & {32.6} & 44.0\\
    \rowcolor{Gray} LEAD & \cmark &\cmark & \cmark & 85.4 & 85.0 & 86.3 & 90.9 & 86.2 & 93.1 & \textbf{87.8} & 59.9 & 46.1 & 51.3 & 45.0 & 45.9 & 56.3 & \textbf{50.8} & \textbf{76.6} \\
    \hline
    UMAD~\cite{liang2021_umad} & \xmark & \cmark & \cmark & 79.1  & 77.4  & 87.4  & 90.7  & 90.4  & 97.2  & 87.0   & 59.0 & {44.3} &  {50.1} & {42.1} & {32.0} & {55.3} & {47.1} & 58.3 \\
    \rowcolor{Gray} + LEAD & \cmark &\cmark &\cmark & 88.7  & 89.2 & 90.8 & 95.8 & 90.4 & 97.1 & \textbf{92.0} {\footnotesize \textcolor{darkgreen}{(+5.0)}} & 59.3 & 46.0 & 49.8 & 43.8 & 43.9 & 56.0 & \textbf{49.8} {\footnotesize \textcolor{darkgreen}{(+2.7)}} & \textbf{67.2} {\footnotesize \textcolor{darkgreen}{(+8.9)}}\\
    \hline
    GLC~\cite{qu2023_glc} & \cmark  & \cmark & \xmark & {81.5}      & 84.5     & 89.8    & 90.4      &88.4    & 92.3      & 87.8         & 63.3  & 50.5  & 54.9    & 50.9   & 49.6    & 61.3     & {55.1} & 73.1\\
    \rowcolor{Gray} + LEAD & \cmark &\cmark & \xmark& 83.0 & 86.8 & 87.6 & 92.0 & 87.1 & 93.2 & \textbf{88.3} {\footnotesize \textcolor{darkgreen}{(+0.5)}}  & 63.5 & 50.7 & 55.1 & 51.6 & 49.9 & 61.3 & \textbf{55.4} {\footnotesize \textcolor{darkgreen}{(+0.3)}} &\textbf{76.8} {\footnotesize \textcolor{darkgreen}{(+3.7)}}\\
    \bottomrule
    \end{tabular}%
  }
  \label{tab:opda_rest}%
  \vspace{-0.22in}
\end{table*}%

\begin{table*}[t]
  \centering
  \vspace{-0.00in}
  \caption{H-score (\%) comparison in OSDA scenario on Office-Home, Office-31, and VisDA.}
   \vspace{-0.10in}
  \addtolength{\tabcolsep}{-4.5pt}
  \resizebox{0.99\textwidth}{!}{
    \begin{tabular}{lccc|ccccccccccccc|ccccccc|c}
    \toprule
    \multirow{2}[1]{*}{Methods} & \multirow{2}[1]{*}{U} & \multirow{2}[1]{*}{SF} & \multirow{2}[1]{*}{CF} & \multicolumn{13}{c|}{Office-Home}                                                           & \multicolumn{7}{c|}{Office-31}  &
    \multicolumn{1}{c}{VisDA} \\
\cline{5-25}  & & & & Ar2Cl & Ar2Pr & Ar2Re & Cl2Ar & Cl2Pr & Cl2Re & Pr2Ar & Pr2Cl & Pr2Re & Re2Ar & Re2Cl & Re2Pr & \textbf{\ Avg\ }   & A2D & A2W & D2A & D2W & W2A  & W2D  & \textbf{Avg}   & \textbf{S2R} \\
    \midrule
    CMU~\cite{fu2020_cmu} & \xmark  & \xmark & \cmark   & 55.0  & 57.0  & 59.0  & 59.3  & 58.2  & 60.6  & 59.2  & 51.3  & 61.2  & 61.9  & 53.5  & 55.3  & 57.6 & 52.6 & 55.7 & 76.5 & 75.9 & 65.8 & 64.7  & 65.2  & 54.2 \\
    DANCE~\cite{saito2020_unida_dance} & \cmark & \xmark & \cmark & 6.5   & 9.0   & 9.9   & 20.4  & 10.4  & 9.2   & 28.4  & 12.8  & 12.6  & 14.2  & 7.9   & 13.2  & 12.9 & 84.9 & 78.8 & 79.1 & 78.8 & 68.3 & 78.8 & 79.8  & 67.5 \\
    DCC~\cite{li2021_dcc} & \cmark  & \xmark  & \xmark  & 56.1  & 67.5  & 66.7  & 49.6  & 66.5  & 64.0  & 55.8  & 53.0  & 70.5  & 61.6  & 57.2  & 71.9  & 61.7 & 58.3 & 54.8 & 67.2 & 89.4 & 85.3 & 80.9 & 72.7  & 59.6 \\
    OVANet~\cite{saito2021_ovanet} & \xmark & \xmark & \cmark   & 58.6  & 66.3  & 69.9  & 62.0  & 65.2  & 68.6  & 59.8  & 53.4  & 69.3  & 68.7  & 59.6  & 66.7  & 64.0 & 90.5 & 88.3 & 86.7 & 98.2 & 88.3 & 98.4  & 91.7 & 66.1 \\
    GATE~\cite{chen2022_gate} & \cmark & \xmark & \cmark  & 63.8  & 70.5  & 75.8  & 66.4  & 67.9  & 71.7  & 67.3 & 61.5  & 76.0  & 70.4  & 61.8 & 75.1  & 69.0  & 88.4 & 86.5 & 84.2 & 95.0 & 86.1 & 96.7 & 89.5  & 70.8 \\
    \midrule
    \midrule
    Source-only &- &- &-    & 46.1  & 63.3  & 72.9  & 42.8  & 54.0  & 58.7  & 47.8  & {36.1} & 66.2  & 60.8  & 45.3  & 68.2  & 55.2  & 78.2 & 72.1 & 44.2 & 82.2 & 52.1 & 88.8 & 69.6      &29.1  \\
    SHOT-O~\cite{liang2020_shot} & \xmark & \cmark & \cmark & 37.7  & 41.8  & 48.4  & 56.4  & 39.8  & 40.9  & 60.0  & 41.5  & 49.7  & 61.8  & 41.4  & 43.6  & 46.9  & 80.2 & 71.6 & 64.3 & 93.1 & 64.0 & 91.8 & 77.5  & 28.1 \\
    \rowcolor{Gray} LEAD & \cmark & \cmark & \cmark & 60.7 & 70.8 & 76.5 & 61.0 & 68.6 & 70.8 & 65.5 & 59.8 & 74.2 & 64.8 & 57.7 & 75.8 & \textbf{67.2} & 84.9 & 85.1 & 90.2 & 94.8 & {90.3} & 96.5 & \textbf{90.3} & \textbf{74.2}\\
    \hline
    UMAD~\cite{liang2021_umad} & \xmark  & \cmark &\cmark  & 59.2  & {71.8}  & 76.6  & 63.5  & 69.0  & 71.9  & 62.5  & 54.6  & 72.8  & 66.5  & 57.9  & 70.7  & 66.4 & 88.5 & 84.4 & 86.8 & 95.0 & 88.2 & 95.9  & 89.8  & 66.8 \\
    \rowcolor{Gray} + LEAD & \cmark & \cmark & \cmark & 62.6 & 73.6 & 77.3 & 61.3 & 66.2 & 70.7 & 63.3 & 54.6 & 75.1 & 66.7 & 63.5 & 75.6 & \textbf{67.6} & 90.4 & 89.6 & 90.3 & 95.4 & 90.7 & 97.3 & \textbf{92.3} & \textbf{70.2} \\
    \rowcolor{Gray} & & & & & & & & & & & & & & & &{\small \textcolor{darkgreen}{+1.2}} & & & & & & &{\small \textcolor{darkgreen}{+2.5}} & {\small \textcolor{darkgreen}{+3.4}}\\
    \hline
    GLC~\cite{qu2023_glc} & \cmark  & \cmark &\xmark & 65.3    & 74.2     & 79.0       & 60.4       & 71.6      & 74.7     & 63.7     & 63.2      & 75.8     & 67.1      & 64.3     & 77.8 & {69.8}   & 82.6 & 74.6 & 92.6 & 96.0 & 91.8 & 96.1  & {89.0}      & 72.5  \\
    \rowcolor{Gray} + LEAD & \cmark & \cmark &\xmark & 64.9 & 74.1 & 78.7 & 60.7 & 71.1 & 74.4 & 65.2 & 64.2 & 76.2 & 67.5 & 62.6 & 80.1 & \textbf{70.0} & 84.7 & 82.7 & 91.0 & 94.9 & 90.8 & 96.3 & \textbf{90.1} & \textbf{73.1} \\
    \rowcolor{Gray} & & & & & & & & & & & & & & & &{\small \textcolor{darkgreen}{+0.2}} & & & & & & &{\small \textcolor{darkgreen}{+1.1}} & {\small \textcolor{darkgreen}{+0.6}}\\
    \bottomrule
    \end{tabular}%
  }
   \vspace{-0.10in}
  \label{tab:osda}%
\end{table*}%

\begin{table*}[t]
  \centering
  \vspace{-0.00in}
  \caption{ Accuracy comparison (\%) in PDA scenario on Office-Home, Office-31, and VisDA.}
   \vspace{-0.10in}
  \addtolength{\tabcolsep}{-4.5pt}
  \resizebox{0.99\textwidth}{!}{
    \begin{tabular}{lccc|ccccccccccccc|ccccccc|c}
    \toprule
    \multirow{2}[1]{*}{Methods} & \multirow{2}[1]{*}{U} & \multirow{2}[1]{*}{SF} & \multirow{2}[1]{*}{CF} & \multicolumn{13}{c|}{Office-Home}                                                           & \multicolumn{7}{c|}{Office-31}  &
    \multicolumn{1}{c}{VisDA} \\
\cline{5-25}  & & & & Ar2Cl & Ar2Pr & Ar2Re & Cl2Ar & Cl2Pr & Cl2Re & Pr2Ar & Pr2Cl & Pr2Re & Re2Ar & Re2Cl & Re2Pr & \textbf{\ Avg\ }   & A2D & A2W & D2A & D2W & W2A  & W2D  & \textbf{Avg}   & \textbf{S2R} \\
    \midrule
    CMU~\cite{fu2020_cmu} & \xmark  & \xmark & \cmark & 50.9  & 74.2  & 78.4  & 62.2  & 64.1  & 72.5  & 63.5  & 47.9  & 78.3  & 72.4  & 54.7  & 78.9  & 66.5 & 84.1 & 84.2  &69.2 & 97.2 & 66.8 & 98.8 & 83.4  &65.5\\
    DANCE~\cite{saito2020_unida_dance} & \cmark & \xmark & \cmark & 53.6  & 73.2  & 84.9  & 70.8  & 67.3  & 82.6  & 70.0  & 50.9  & 84.8  & 77.0  & 55.9  & 81.8  & 71.1 & 77.1 & 71.2 & 83.7 & 94.6 & 92.6 & 96.8 &86.0 & 73.7 \\
    DCC~\cite{li2021_dcc}  & \cmark & \xmark & \xmark & 54.2  & 47.5  & 57.5  & 83.8  & 71.6  & 86.2  & 63.7  & 65.0 & 75.2  & 85.5 & 78.2  & 82.6  & 70.9 & 87.3 & 81.3 & 95.4 & 100.0 & 95.5 & 100.0 & 93.3 & 72.4\\
    OVANet~\cite{saito2021_ovanet} & \xmark & \xmark & \cmark & 34.1  & 54.6  & 72.1  & 42.4  & 47.3  & 55.9  & 38.2  & 26.2  & 61.7  & 56.7  & 35.8  & 68.9  & 49.5 & 69.4 & 61.7 & 61.4 & 90.2 & 66.4 & 98.7 & 74.6 & 34.3 \\
    GATE~\cite{chen2022_gate} & \cmark & \xmark & \cmark & 55.8  & 75.9  & 85.3  & 73.6  & 70.2  & 83.0  & 72.1  & 59.5  & 84.7  & 79.6  & 63.9  & 83.8  & 74.0 & 89.5 & 86.2 & 93.5 & 100.0 & 94.4 & 98.6 & 93.7 & 75.6 \\
    \midrule
    \midrule
    Source-only &- &- &- & 45.9  & 69.2  & 81.1  & 55.7  & 61.2  & 64.8  & 60.7  & 41.1  & 75.8  & 70.5  & 49.9  & 78.4  & 62.9   & 90.4 & 79.3 & 79.3 & 95.9 & 84.3 & 98.1 & 87.8 & 42.8  \\
    SHOT-P~\cite{liang2020_shot} & \xmark & \cmark & \cmark & 64.7  & 85.1  & 90.1  & 75.1  & 73.9 & 84.2  & 76.4  & 64.1  & 90.3  & 80.7  & 63.3  & 85.5  & 77.8 & 89.8 & 84.4 & 92.2 & 96.6 & 92.2 & 99.4 & 92.4 & 74.2\\
    \rowcolor{Gray} LEAD & \cmark &\cmark & \cmark & 58.2 & 83.1 & 87.0 & 70.5 & 75.4 & 83.3 & 73.7 & 50.4 & 83.7 & 78.3 & 58.7 & 83.2 & \textbf{73.8} & 89.8 & 93.9 & 95.6 & 98.6 & 96.0 & 99.4 & \textbf{95.5}  & \textbf{75.3}\\
    \hline
    UMAD~\cite{liang2021_umad} & \xmark & \cmark &\cmark & 51.2     & 66.5     & 79.2     & 63.1     & 62.9     & 68.2     & 63.3     & 56.4     & 75.9     & 74.5     & 55.9     & 78.3     & 66.3  & 85.4 & 85.1 & 83.5 & 97.6 & 86.2 & 99.4 & 89.5 & 68.5 \\
    \rowcolor{Gray} + LEAD & \cmark &\cmark &\cmark & 61.1 & 78.2 & 87.5 & 70.5 & 78.5 & 81.8 & 74.1 & 60.4 & 83.3 & 78.8 & 63.2 & 82.1 & \textbf{75.0} & 89.2 & 90.8 & 90.6 & 98.0 & 93.0 & 100.0 & \textbf{93.6} & \textbf{78.1}\\
    \rowcolor{Gray} & & & & & & & & & & & & & & & &{\small \textcolor{darkgreen}{+8.7}} & & & & & & &{\small \textcolor{darkgreen}{+4.1}} & {\small \textcolor{darkgreen}{+9.6}}\\
    \hline
    GLC~\cite{qu2023_glc} & \cmark  & \cmark  & \xmark & 55.9      & 79.0      & 87.5      & 72.5      & 71.8      & 82.7      & 74.9      & 41.7      & 82.4      & 77.3      & 60.4      & 84.3      & 72.5 & 89.8 & 89.8 & 92.8 & 96.6 & 96.1 & 99.4 & 94.1 & 76.2\\
    \rowcolor{Gray} + LEAD & \cmark & \cmark  & \xmark  & 56.6 & 81.2 & 88.1 & 72.5 & 72.2 & 83.2 & 75.3 & 43.6 & 83.4 & 78.2 & 59.7 & 84.2 & \textbf{73.2} & 89.8 & 93.2 & 95.6 & 96.9 & 96.0 & 99.4 & \textbf{95.1} & \textbf{75.5} \\
    \rowcolor{Gray} & & & & & & & & & & & & & & & &{\small \textcolor{darkgreen}{+0.7}} & & & & & & &{\small \textcolor{darkgreen}{+1.0}} & {\small \textcolor{white}{-0.7}}\\
    \bottomrule
    \end{tabular}%
  }
   \vspace{-0.20in}
  \label{tab:pda}%
\end{table*}%

\vspace{-0.10in}
\section{Experiments}
\subsection{Experimental Setup}
\noindent \textbf{Dataset.} We empirically verify the effectiveness of our LEAD on four datasets, including Office-31~\cite{office31}, Office-Home~\cite{officehome}, VisDA~\cite{visda}, and DomainNet~\cite{domainnet}. For a fair comparison, we evaluate LEAD on partial-set domain adaptation (PDA), open-set domain adaptation (OSDA), and open-partial-set domain adaptation (OPDA). Details of classes split are summarized in Table~\ref{tab:label_split}.
\\
\noindent \textbf{Evaluation.} We utilize the same evaluation metric as previous works. In PDA scenarios, we report the classification accuracy for evaluation. In OSDA and OPDA scenarios, we report the H-score for performance comparison, which is a harmonic mean of accuracy over common data and accuracy over private data. It is worth noting that LEAD can be incorporated effortlessly into most existing SF-UniDA methods, for potential performance improvement. To demonstrate this merit, we integrate LEAD with two representative methods: UMAD~\cite{liang2021_umad} and GLC~\cite{qu2023_glc}. 
\\ 
\noindent \textbf{Implementation Details.} All the experiments are conducted on a single RTX-3090 GPU with PyTorch-1.10. We adopt the same network architecture as existing baseline methods. During target model adaptation, we apply the SGD optimizer with momentum 0.9. The batch size is set to 64 for all experiments. The learning rate is set to 1e-3 for Office-31 and Office-Home, and 1e-4 for VisDA and DomainNet. As for $\lambda$, it is set to 0.3 for Office-31, 1.0 for VisDA, and 2.0 for Office-Home and DomainNet. During inference, we adopt the same protocol like~\cite{qu2023_glc}. More details are presented in the supplementary.
\subsection{Experimental Results}
\vspace{-0.2in}
\noindent \textbf{Results for OPDA Scenarios.} To verify the effectiveness of LEAD, we first conduct experiments in the most challenging OPDA scenarios, where both source and target domains contain private data. Table~\ref{tab:opda_officehome} summarizes the results on Office-Home, and Table~\ref{tab:opda_rest} details the results on Office-31, DomainNet, and VisDA. These results demonstrate that LEAD achieves performance surpassing or comparable to existing methods based on K-means clustering for segregating common and private data. In comparison with methods without clustering, LEAD achieves significantly superior performance. For instance, on Office-Home, DomainNet, and VisDA datasets, LEAD outperforms UMAD by 4.9\%, 3.7\%, and 18.3\%, respectively. We attribute this to our instance-level adaptive decision strategy, which effectively circumvents the limitations imposed by a fixed threshold. While GLC achieves commendable performance on its own, the integration with LEAD further amplifies its effectiveness. For example, LEAD improves GLC in the H-score on VisDA from 73.1\% to 76.8\%. The integration with LEAD proves even more significant against UMAD, with the H-score on Office-Home and VisDA boosting from 70.1\%/58.3\% to 78.0\%/67.2\%, respectively.
\\
\noindent \textbf{Results for OSDA Scenarios.} We then conduct experiments in OSDA scenarios, where only the target domain involves private data. Table~\ref{tab:osda} lists the results, demonstrating the superiority of LEAD against baselines. Specifically, LEAD obtains the H-score of 67.2\%, 90.3\%, and 74.2\% on Office-Home, Office-31, and VisDA datasets, respectively. Similar to the observations in OPDA scenarios, LEAD also exhibits performance improvement to existing methods. LEAD improves the H-score of UMAD by 1.2\%, 2.5\%, and 3.4\% across the Office-Home, Office-31, and VisDA datasets, respectively. The results again verify the merit of LEAD when incorporated with existing methods.
\\
\noindent \textbf{Results for PDA Scenarios.} 
Finally, we verify the effectiveness of LEAD in PDA scenarios, where the target domain's label sets are subsets of those in the source domain. Results presented in Table~\ref{tab:pda} show that LEAD achieves superior performance even compared to methods tailored for PDA. In particular, LEAD attains the overall accuracy of 73.8\%, 95.5\%, and 75.3\% on Office-Home, Office-31, and VisDA datasets, individually. LEAD contributes the accuracy increase of 8.7\%, 4.1\%, and 9.6\% to UMAD on these datasets. 
Note that, there is a slight degradation (-0.7\%) for GLC w/ LEAD. It may arise from gradient conflict, causing the optimization process to inadvertently compromise one objective in favor of improving the overall objective. We will pursue better integration strategies in future work.

\begin{table}[t]
    \centering
    \caption{Ablation studies of LEAD in OPDA scenarios.}
    \vspace{-0.10in}
    \addtolength{\tabcolsep}{4.0pt}
    \resizebox{0.99\linewidth}{!}{
    \begin{tabular}{ccc|ccc}
        \toprule
        $\mathcal{L}_{ce}$ & $\mathcal{L}_{reg}$ & $\mathcal{L}_{con}$ & Office-31 & Office-Home & VisDA \\
        \midrule
        - & -  & - & 64.9 & 60.9 & 25.7\\
        \cmark & - & - & 82.0 & 73.6 & 61.5\\
        \cmark & \cmark & - & 84.5 & 74.5 & 67.3 \\
        \cmark & - & \cmark & 87.2 & 74.5 & 75.7 \\
        \cmark & \cmark & \cmark & 87.8 & 75.0 & 76.6 \\
        \bottomrule
    \end{tabular}
    }
    \vspace{-0.25in}
    \label{tab: ablation}
\end{table}

\begin{figure*}[ht]
    \centering
    \vspace{-0.05in}
    \includegraphics[width=0.90\textwidth]{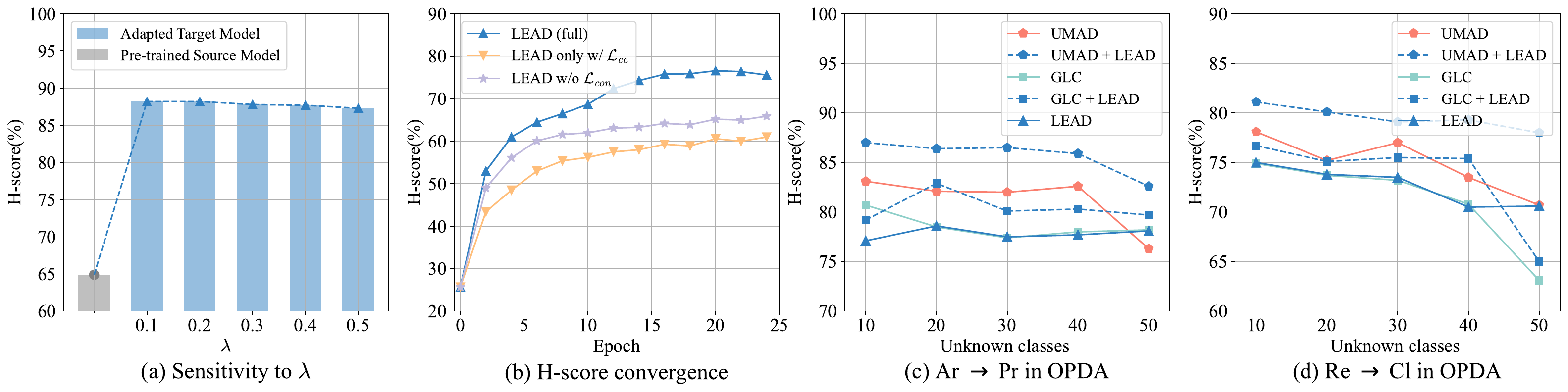}
    \vspace{-0.15in}
    \caption{\textbf{Robustness analysis}. (a) shows the sensitivity to $\lambda$ on Office-31. (b) presents the H-score curves on VisDA. (c-d) present robustness analysis when varying the unknown private categories. 
    }
    \vspace{-0.25in}
    \label{fig: analysis_1}
\end{figure*}

\subsection{Experimental Analysis}
\vspace{-0.1in}
\noindent \textbf{Ablation Study.} To assess the contribution of different components within LEAD, we carry out extensive ablation studies in OPDA scenarios. The results, detailed in Table~\ref{tab: ablation}, confirm that each component is not only effective but also complementary to the others.
\\
\noindent \textbf{Efficiency Comparison.}
As previously stated, LEAD employs the distance to target prototypes and source anchors for adaptive pseudo-labeling. This design, in contrast to clustering-based methods, notably reduces the need for computational resources. Table~\ref{tab: runtime} juxtaposes the runtime of GLC~\cite{qu2023_glc} and LEAD in OPDA scenarios. It is clear that LEAD requires minimal resources for deriving decision boundaries, whereas GLC's resource demands escalate dramatically with data scale. In particular, on DomainNet, with about 0.6 million images across 345 categories, GLC averages 897.65 seconds, while LEAD takes only 0.29 seconds on the same platform. Moreover, LEAD also shows substantial resource savings even on normal-scale datasets, e.g., on VisDA with over 75\% reduction.
\begin{table}[t]
    \centering
    \vspace{-0.00in}
    \caption{Runtime (s) to derive pseudo-label decision boundaries.}
    \vspace{-0.10in}
    \addtolength{\tabcolsep}{5.0pt}
    \resizebox{0.99\linewidth}{!}{
    \begin{tabular}{l|cccc}
        \toprule
         & Office-31 & Office-Home & VisDA & DomainNet\\
        \midrule
        GLC~\cite{qu2023_glc} & 0.19 & 0.51 & 0.46 & 897.65\\
        LEAD & 0.02 & 0.02 & 0.11 & 0.29\\
        \bottomrule
    \end{tabular}
    }
    \vspace{-0.30in}
    \label{tab: runtime}
\end{table}

\noindent \textbf{Parameter Sensitivity.} We investigate the parameter sensitivity of $\lambda$ on Office-31 in OPDA scenarios, where the range of $\lambda$ is set to $[0.1, 0.2, 0.3, 0.4, 0.5]$. Figure~\ref{fig: analysis_1} (a) illustrates that the results are stable around the selected parameter $\lambda = 0.3$. Through oracle validation, we may find the optimal parameter configuration for this setting is $\lambda = 0.1$.

\noindent \textbf{Training Stability.} Figure~\ref{fig: analysis_1} (b) visualizes the training curves of our LEAD on VisDA in OPDA scenarios. 
Generally, the training procedure of LEAD is stable and effective.

\noindent \textbf{Varying Unknown Classes.} 
With the rise in unknown private classes, accurately distinguishing common from private data becomes more challenging. To examine the robustness of our LEAD, we compared it with GLC and UMAD in OPDA scenarios using the Office-Home dataset. The results in Figure~\ref{fig: analysis_1} (c-d) show that LEAD achieves stable results and consistently boosts existing methods.

\begin{figure}[t]
    \centering
    \vspace{-0.12in}
    \includegraphics[width=0.45\textwidth]{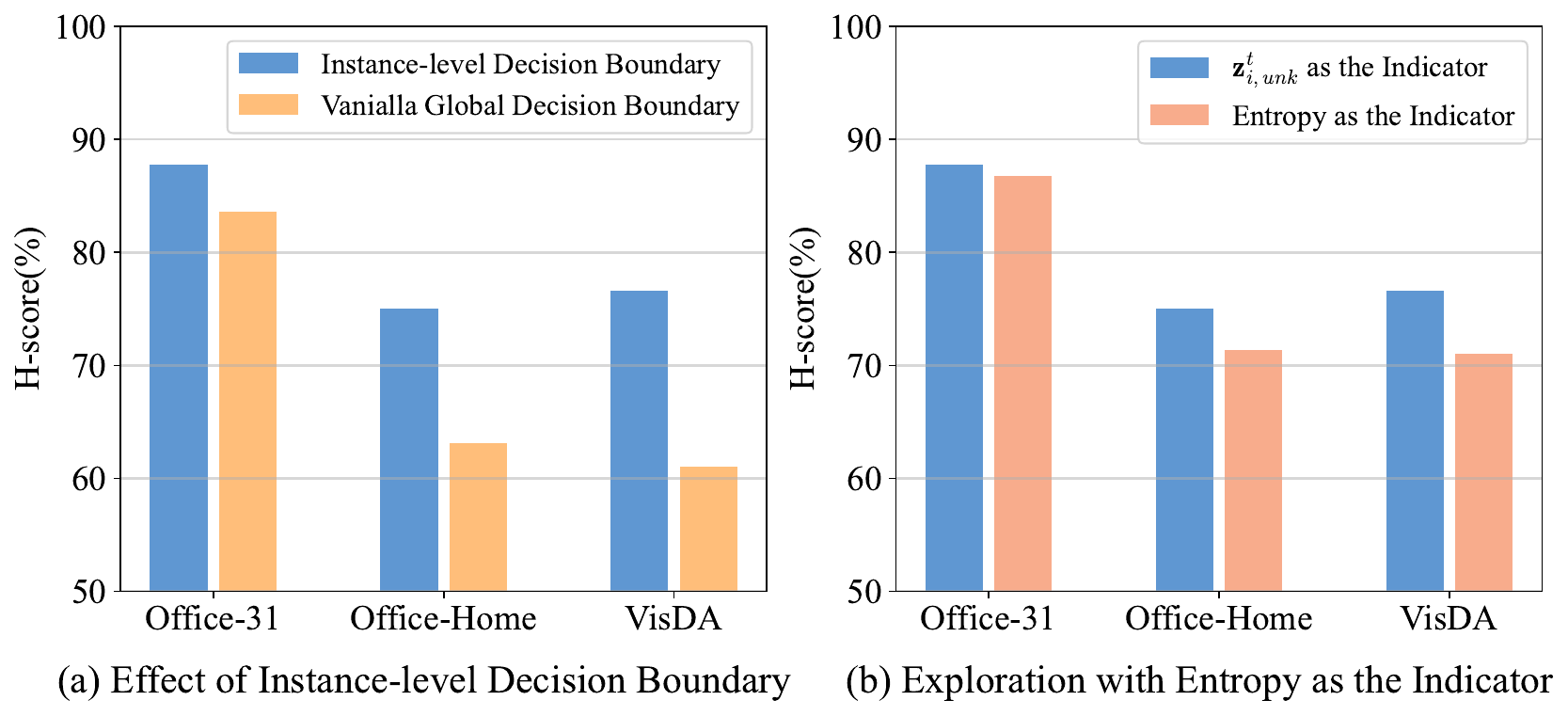}
    \vspace{-0.13in}
    \caption{\textbf{Methodology analysis.} (a) compares the effectiveness of our instance-level decision strategy against the vanilla global decision strategy. (b) examines the efficacy of using Entropy as the indicator for private data.
    }
    \vspace{-0.30in}
    \label{fig: analysis_2}
\end{figure}

\noindent \textbf{Effect of Instance-level Decision Boundary.} To evaluate the contribution of our instance-level pseudo-labeling strategy, we conduct a comparison against the intuitive design, which uses the average of $\mu_{com}$ and $\mu_{pri}$ as the threshold. As depicted in Figure~\ref{fig: analysis_2} (a), instance-level decision boundary strategy significantly outperforms the vanilla global decision design (75.0\% vs 63.1\% on Office-Home in OPDA scenarios). We attribute this to that vanilla design discounts the incosistency of covariate shifts across categories, and also the variability among samples from the same category.

\noindent \textbf{Exploration with Entropy as the Indicator.} Contrasting with existing threshold-based methods that utilize Entropy as the indicator for differentiating common and private data, we propose decomposing the feature and using the projection $\mathbf{z}^t_{i, unk}$ on the source-unknown space as an alternative indicator. To empirically substantiate the superiority, we remold our framework to incorporate Entropy as the indicator. Figure~\ref{fig: analysis_2} (b) depicts the results. An observation is that while employing entropy as the indicator within our framework is feasible, its performance is inferior compared to $\mathbf{z}^t_{i, unk}$. We speculate that this inferiority might result from entropy's limited capacity to distinguish between common samples with high uncertainty and private unknown samples.

\vspace{-0.05in}
\section{Conclusion}
\par In this paper, we delve into source-free universal domain adaptation (SF-UniDA). Different from existing methods that leverage hand-crafted thresholding strategies or feature-clustering algorithms, we propose a novel idea of LEArning Decomposition (LEAD) tailored for SF-UniDA. Technically, LEAD first decomposes features into source-known and -unknown two parts, then establishes instance-level decision boundaries for pseudo-labeling. It offers an elegant view to identify target-private unknown data without tedious tuning a threshold or relying on unstable clustering. Extensive experiments across various scenarios have verified the effectiveness of our LEAD. Besides, it is appealing in view that LEAD can be seamlessly incorporated into existing methods to further boost performance. Remarkably, LEAD improves UMAD by 7.9\% overall H-score in the OPDA scenario on Office-Home dataset.
\\
\noindent\textbf{Acknowledgment:} This work is supported by the National Natural Science Foundation of China (No. 62372329), in part by the National Key Research and Development Program of China (No. 2021YFB2501104), in part by Shanghai Rising Star Program (No.21QC1400900), in part by Tongji-Qomolo Autonomous Driving Commercial Vehicle Joint Lab Project, and in part by Xiaomi Young Talents Program, and in part by the International Exchange Program for Graduate Students, Tongji University (No.2023020053).

\maketitlesupplementary
\appendix
\section{More Related Work}
\vspace{-0.05in}
\noindent \textbf{Source-free Domain Adaptation (SFDA).} SFDA aims to adapt a pre-trained source model to a new domain without needing access to the original source data. Existing methods in this field generally fall into two categories: source distribution estimation methods~\cite{li2020_3cgan, nayak2021_DI, zhang2022_dac, tian2021_vdm} and self-training methods~\cite{liang2020_shot, yang2021_nrc, wang2021_ATP, sun2022_kuda}. The former often utilizes generative networks to create synthetic-labeled data or to transfer target data into the style of the source domain, thereby bridging the domain gap. The latter, drawing inspiration from semi-supervised learning, typically adopts various pseudo-labeling techniques for model adaptation. However, these methods mainly focus on the vanilla closed-set scenarios where the source and target domains share identical label spaces, significantly limiting their applicability. In contrast, this paper focuses on Source-free Universal Domain Adaptation (SF-UniDA), targeting the more challenging setting that encompasses both covariate and label shifts.

\noindent \textbf{Out-of-distribution (OOD) Detection.} The primary objective of OOD detection is to identify test (target) samples that are distinct from the training (source) distribution. Typically, in the literature of OOD, the distribution is referred to as `label distribution', with OOD samples being those unrecognizable or exclusive to the source label space. As such, OOD detection can be considered a component of UniDA/SF-UniDA, both of which aim to reject these target-private unknown data from the target domain. In addressing OOD, prevailing techniques~\cite{hendrycks2016_oodbaseline, hsu2020_odin, wei2022_mitigating} often use metrics such as the maximum of softmax outputs or confidence scores. Recent studies~\cite{yang2021_semantically, ming2022_poem, katz2022_training} have begun incorporating a collection of OOD samples for outlier exposure during model pre-training, fostering a clearer distinction between in-distribution (ID) and OOD samples. However, these methods typically depend on manually set thresholds for identifying target-private data, which can be both tedious and sub-optimal. In contrast, we start from the perspective of feature decomposition and utilize the feature projection onto source-unknown space as the indicator. Then, we consider the distance to both target prototypes and source anchors to establish adaptive instance-level decision boundaries. Our solution is effective under conditions of both covariate and label shift, offering a more flexible and robust approach for UniDA/SF-UniDA.

\vspace{-0.05in}
\section{More Details about Methodology}
\subsection{Target Prototype Construction}
\par In the preceding discussion, we outlined the construction of instance-level pseudo-labeling decision boundaries. These boundaries are defined based on the distance of samples to source anchors, denoted as $\{\mathbf{c}^s_c \in \mathbb{R}^D |{c=1,\dots, C}\}$, and target prototypes, denoted as $\{\mathbf{c}^t_c \in \mathbb{R}^D |{c=1,\dots, C}\}$. Source anchors are derived directly from the classifier weight $W_{cls}$, where $W_{cls} \in \mathbb{R}^{C\times D}$. Regarding the target prototypes, we employ a top-$K$ sampling strategy for their construction. Specifically, for each category $c$, we identify the top-$K$ instances in the target domain with the highest $\delta_c(f^t_\theta(\mathbf{x}_i^t))$ scores. These instances are then averaged to form the target prototype $\mathbf{c}^t_c$ for category $c$. Formally, 
\begin{equation}
    \begin{aligned}
    \mathcal{M}_c &= \mathop{\arg \max}_{ \mathbf{x}_i^t \in \mathcal{X}_t; \ |\mathcal{M}_c| = K}\  \delta_c(f^t_\theta(\mathbf{x}_i^t)),\\
    \mathbf{c}^t_c &= \frac{1}{K}\sum_{i\in \mathcal{M}_c} g_{\theta}^t(\mathbf{x}^t_i).
    \end{aligned}
\end{equation}
where $\delta_c(f^t_\theta(\mathbf{x}_i^t))$ denotes the $c$-th soft-max probability for the instance $\mathbf{x}_i^t$. As previously mentioned, $f^t_\theta$ and $g^t_\theta$ represent the entire target model and its feature extractor, respectively. A critical challenge is determining the appropriate value for $K$. Following previous work~\cite{qu2023_glc, liu2023_coca}, we adopt a straightforward yet effective method, setting $K = N_t/\hat{C}_t$. Here, $N_t$ signifies the amount of target data, while $\hat{C}_t$ estimates the number of target categories. For the estimation of $\hat{C}_t$, we utilize the Silhouette metric\cite{silhouettes}. This process involves initially enumerating possible values for ${C}_t$, followed by partitioning the target domain into respective clusters using an algorithm such as K-means. Subsequently, the Silhouette metric aids in selecting the most suitable value for $\hat{C}_t$.
Formally, the Silhouette score for $\mathbf{x}_i^t$ is defined as:
\begin{equation}
    \begin{aligned}
        a(\mathbf{x}_i^t) &= \frac{1}{|\mathcal{C_I}| - 1}\sum_{\mathbf{x}_i^t\in \mathcal{C_I}, i\neq j} d(g_{\theta}^t(\mathbf{x}_i^t), g_{\theta}^t(\mathbf{x}_j^t)),\\
        b(\mathbf{x}_i^t) &= \mathop{\min}_{\mathcal{J\neq I}} \frac{1}{|\mathcal{C_J}|} \sum_{\mathbf{x}_j^t\in \mathcal{C_J}} d(g_{\theta}^t(\mathbf{x}_i^t), g_{\theta}^t(\mathbf{x}_j^t)),\\
        s(\mathbf{x}_i^t) &= \frac{b(\mathbf{x}_i^t) - a(\mathbf{x}_i^t)}{\mathop{\max} \{a(\mathbf{x}_i^t), b(\mathbf{x}_i^t)\}}.
    \end{aligned}
\end{equation}
where $a(\mathbf{x}_i^t)$ and $b(\mathbf{x}_i^t)$ are functions used to calculate the distance of a data sample $\mathbf{x}_i^t$ to its own cluster $\mathcal{C_I}$ and to other clusters $\mathcal{C_{J, J\neq I}}$, respectively. The term $|\mathcal{C_I}|$ denotes the size of the cluster $\mathcal{C_I}$. $d(,)$ is used to measure the distance between two data samples. The appropriateness of the clustering configuration is assessed by the Silhouette values of the data samples. A majority of high Silhouette values indicate a suitable clustering configuration. Conversely, predominantly low Silhouette values suggest that the clustering configuration may have an inappropriate number of clusters, either too many or too few.

\begin{figure*}[ht]
    \centering
    \vspace{-0.05in}
    \includegraphics[width=0.99\textwidth]{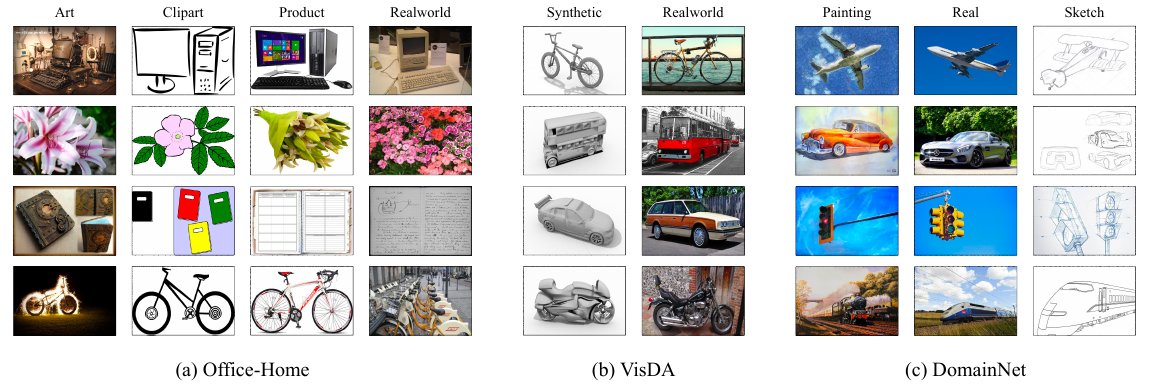}
    \vspace{-0.125in}
    \caption{ Representative examples from the benchmark datasets used in our study, illustrating various types of domain shift. The selected samples highlight the distinct characteristics and environments of the domains within each dataset.
    }
    \vspace{-0.15in}
    \label{fig: dataset}
\end{figure*}

\subsection{Feature Consensus Regularization}
\par Existing literature~\cite{yang2021_gsfda, yang2021_nrc, qu2023_glc} in SFDA and SF-UniDA have observed that the integration of consensus regularization with nearest neighbors in the feature space significantly contributes to stable performance. Building upon these findings, we incorporate the feature consensus learning objective $\mathcal{L}_{con}$ into our proposed LEAD framework. Specifically, 
\begin{equation}
    \begin{aligned}
        l_{i, c}^t &= \frac{1}{|L_i^t|} \sum_{\mathbf{x}_j^t \in L_i^t} \delta_c(f^t_\theta(\mathbf{x}^t_j)),\\
        \mathcal{L}_{con} &= - \frac{1}{N}\sum_{i=1}^{N}\sum_{c=1}^{C} l_{i, c}^t \log \delta_c(f^t_\theta (\mathbf{x}^t_i)).
    \end{aligned}
\end{equation}
where $L_i^t$ represents the set of nearest neighbors for a given target domain data $\mathbf{x}_i^t$ within the feature space. We utilize the cosine similarity function to identify these nearest neighbors. Consistent with~\cite{qu2023_glc}, we empirically determine the size of the nearest neighbor $|L_i^t| = 4$. 

\subsection{Integration into Existing Methods}
\par As elaborated in the main text of our study, the LEAD framework introduces an innovative approach to SF-UniDA through feature decomposition. This unique concept establishes LEAD as a complementary addition to existing methodologies. To validate this merit, we have integrated LEAD with representative methods, specifically UMAD~\cite{liang2021_umad} and GLC~\cite{qu2023_glc}. Given the unsupervised nature of the SF-UniDA task, this integration is achieved by combining the optimization objectives of LEAD with those of the baseline methods. As a result, the integrated optimization objective is presented as follows:
\begin{align}
    \mathcal{L}_{overall} = \gamma\cdot\mathcal{L}_{LEAD} + (1 - \gamma)\cdot\mathcal{L}_{baseline}
\end{align}
where $\gamma$ is a trade-off hyper-parameter, generally set to 0.7.

\begin{figure*}[t]
    \centering
    \includegraphics[width=0.97\textwidth]{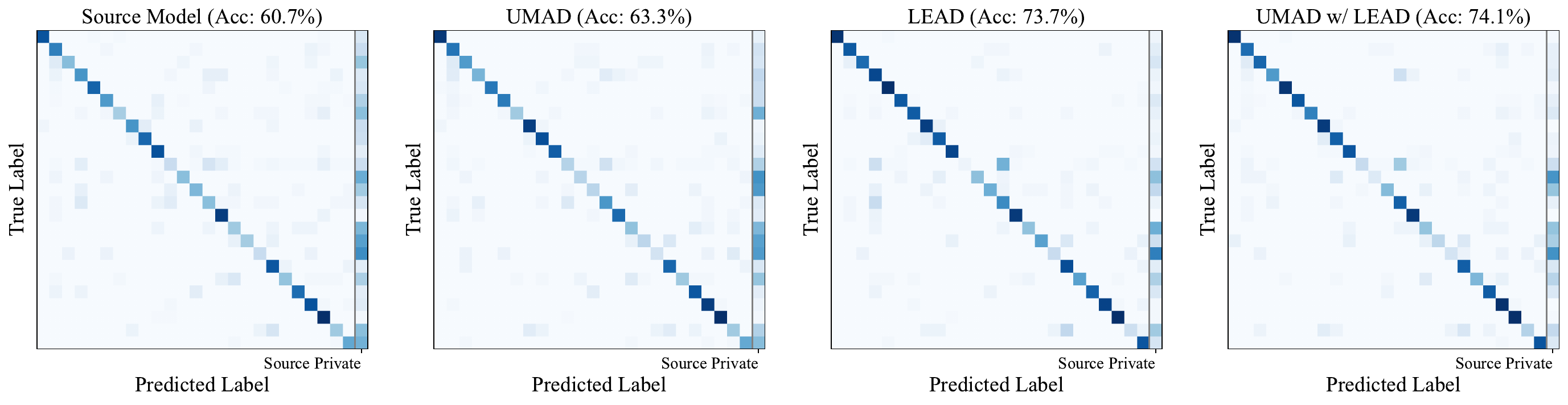}
    \vspace{-0.1in}
    \caption{The confusion matrix for Source Model, UMAD, LEAD, and UMAD w/ LEAD on Pr$\rightarrow$Ar (Office-Home) PDA task.}
    \vspace{-0.05in}
    \label{fig: confu_mat}
\end{figure*}

\definecolor{brightlavender}{rgb}{0.74, 0.65, 0.89}

\begin{figure*}[t]
    \centering
    \vspace{-0.05in}
    \includegraphics[width=0.97\textwidth]{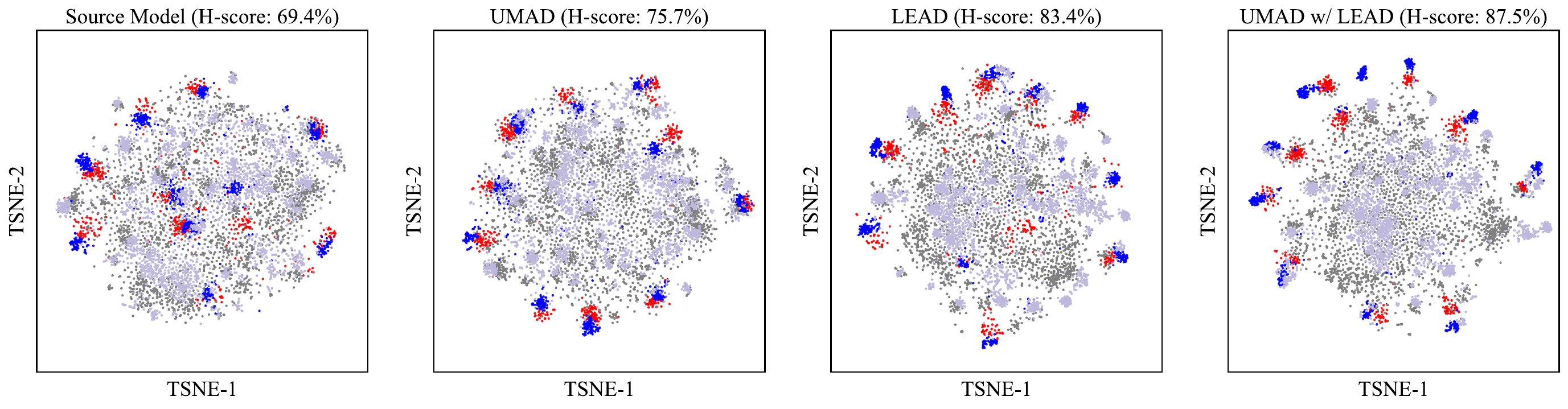}
    \vspace{-0.1in}
    \caption{The t-SNE feature visualization for Source Model, UMAD, LEAD, and UMAD w/ LEAD on Cl$\rightarrow$Re (Office-Home) OPDA task. \textcolor{red}{Points in red} denote unavailable source-common data, \textcolor{gray}{points in gray} represent unavailable source-private data, \textcolor{blue}{points in blue} illustrate target-common data, \textcolor{brightlavender}{points in lavender} illustrate target-private data, individually. It is easy to conclude that both LEAD and UMAD w/ LEAD are effective in achieving a clear separation of common and private data.}
    \vspace{-0.15in}
    \label{fig: tsne}
\end{figure*}

\section{More Details about Experiments}
\subsection{Datasets}
\par In this study, we evaluate the performance and adaptability of our proposed LEAD framework using four standard datasets, each offering distinct challenges. First, Office-31~\cite{office31}, a prevalent domain adaptation benchmark, encompasses 31 object classes (comprising 4,652 images) within an office setting, spread across three distinct domains: DSLR (D), Amazon (A), and Webcam (W). Next, Office-Home~\cite{officehome} presents a more extensive challenge with 65 object classes totaling 15,500 images, categorized into four domains: Artistic images (Ar), Clip-Art images (Cl), Product images (Pr), and Real-world images (Rw). The VisDA~\cite{visda} dataset, a more challenging benchmark, involves 12 object classes; its source domain includes 152,397 synthetic images created from 3D models, and the target domain comprises 55,388 real-world images from Microsoft COCO. Lastly, DomainNet~\cite{domainnet}, the most extensive DA benchmark, contains approximately 0.6 million images across 345 object classes. In line with existing research, our experiments on DomainNet focus on three subsets: Painting (P), Real (R), and Sketch (S). Figure~\ref{fig: dataset} illustrates some representative examples of these benchmarks.

\subsection{Source Model Pre-training}
\par To prepare the source model, we adopt the training recipe~\cite{liang2020_shot, yang2021_nrc, qu2022_bmd, qu2023_glc} widely used in SFDA and SF-UniDA tasks. For a given labeled source domain $\mathcal{D}^s  = \{(\mathbf{x}^s_i, \mathbf{y}^s_i)\}_{i=1}^{N_s}$ where $\mathbf{x}^s_i \in \mathcal{X}^s \subset \mathbb{R}^{X}, \mathbf{y}^s_i \in \mathcal{Y}^s \subset \mathbb{R}^C$. We train the source model $f^s_\theta$ using a smooth cross-entropy loss function, which is detailed as follows:
\begin{equation}
    \mathcal{L}_{src} = -\frac{1}{N}\sum_{i=1}^{N}\sum_{c=1}^{C} q^s_{i,c} \log \delta_c(f^s_\theta(\mathbf{x}^s_i))
\end{equation}
where $\delta_c(f^s_\theta(\mathbf{x}_i^s))$ denotes the $c$-th soft-max probability for the instance $\mathbf{x}_i^s$. $q^s_{i, c}$ corresponds to $c$-th smooth one-hot encoded label for $\mathbf{y}^s_i$, i.e., $q^s_{i, c} = (1 - \beta)\cdot \mymathbb{1}(\mathbf{y}^s_i) + \beta / C$. Here, $\mymathbb{1}$ denotes the one-hot encoding operator, and $\beta$ is the smoothing parameter which is set to 0.1 for all benchmarks. 

\subsection{Target Inference Details}
In the inference phase, we leverage the strategy employed in existing methods~\cite{liu2023_coca, qu2023_glc} to distinguish between common and private data. Specifically, we apply the normalized Shannon Entropy~\cite{shannon_entropy} as the metric for this differentiation. The implementation details of this strategy are as follows:
\begin{align}
    I(\mathbf{x}^t_i) &= - \frac{1}{\log C} \sum_{c=1}^{C} \delta_c(f^t_\theta(\mathbf{x}^t_i)) \log \delta_c(f^t_\theta(\mathbf{x}^t_i))\\
    p(\mathbf{x}^t_i) &= \left\{
\begin{aligned}
 &\text{unknown}, &\text{if $I(\mathbf{x}^t_i) \ge \omega$}\\
 &\text{argmax}(f^t_\theta(\mathbf{x}^t_i)), &\text{if $I(\mathbf{x}^t_i) < \omega$}
\end{aligned}
\right.
\end{align}
where the inference result $p(\mathbf{x}^t_i)$ hinges on the normalized entropy value $I(\mathbf{x}^t_i)$. A higher value of $I(\mathbf{x}^t_i)$ indicates a greater likelihood of the model $f^t_\theta$ classifying the data sample $\mathbf{x}^t_i$ as unknown. In particular, when $I(\mathbf{x}^t_i)$ exceeds a pre-defined threshold $\omega$, the data sample $\mathbf{x}^t_i$ is categorized as target-private. Conversely, if $I(\mathbf{x}^t_i)$ falls below this threshold, the sample is recognized as common. In our experiments across all datasets, we set $\omega = 0.55$, aligning with the thresholds used in existing methods~\cite{qu2023_glc, liu2023_coca}.

\subsection{More Experimental Analysis}
\noindent \textbf{Confusion Matrix Visualization.}
Figure~\ref{fig: confu_mat} provides a visual representation of the confusion matrices for four different models: the model trained solely on source data, UMAD, LEAD, and a combined UMAD with LEAD. This comparison is conducted in the Pr$\rightarrow$Ar (Office-Home) PDA task. The source-only model exhibits a tendency towards inaccurate predictions due to distributional covariate shifts, notably misclassifying target common data as belonging to source private categories. The implementation of model adaptation, as observed in the matrices, evidently mitigates this confusion. However, due to that UMAD is designed primarily for OPDA and OSDA scenarios, its effectiveness is somewhat constrained. In contrast, our LEAD framework demonstrates significant versatility and substantial performance enhancement. Notably, it also proves to be a complementary approach when integrated with UMAD.

\noindent \textbf{t-SNE Feature Visualization.}
In Figure~\ref{fig: tsne}, we provide a t-SNE visualization of the features extracted by the source model, UMAD, LEAD, and UMAD integrated with LEAD. This analysis is conducted within the Cl$\rightarrow$Re (Office-Home) OPDA task. An obvious finding from this visualization is the initially ambiguous boundaries between target private and common data in the feature space, where these data appear intermingled.  As expected, performing model adaptation significantly contributes to the separation between common and private data. Taking a closer look at the visualization, it becomes evident that both LEAD and UMAD w/ LEAD are particularly effective in achieving a more distinct separation between common and private data.

\begin{figure}[t]
    \centering
    \includegraphics[width=0.47\textwidth]{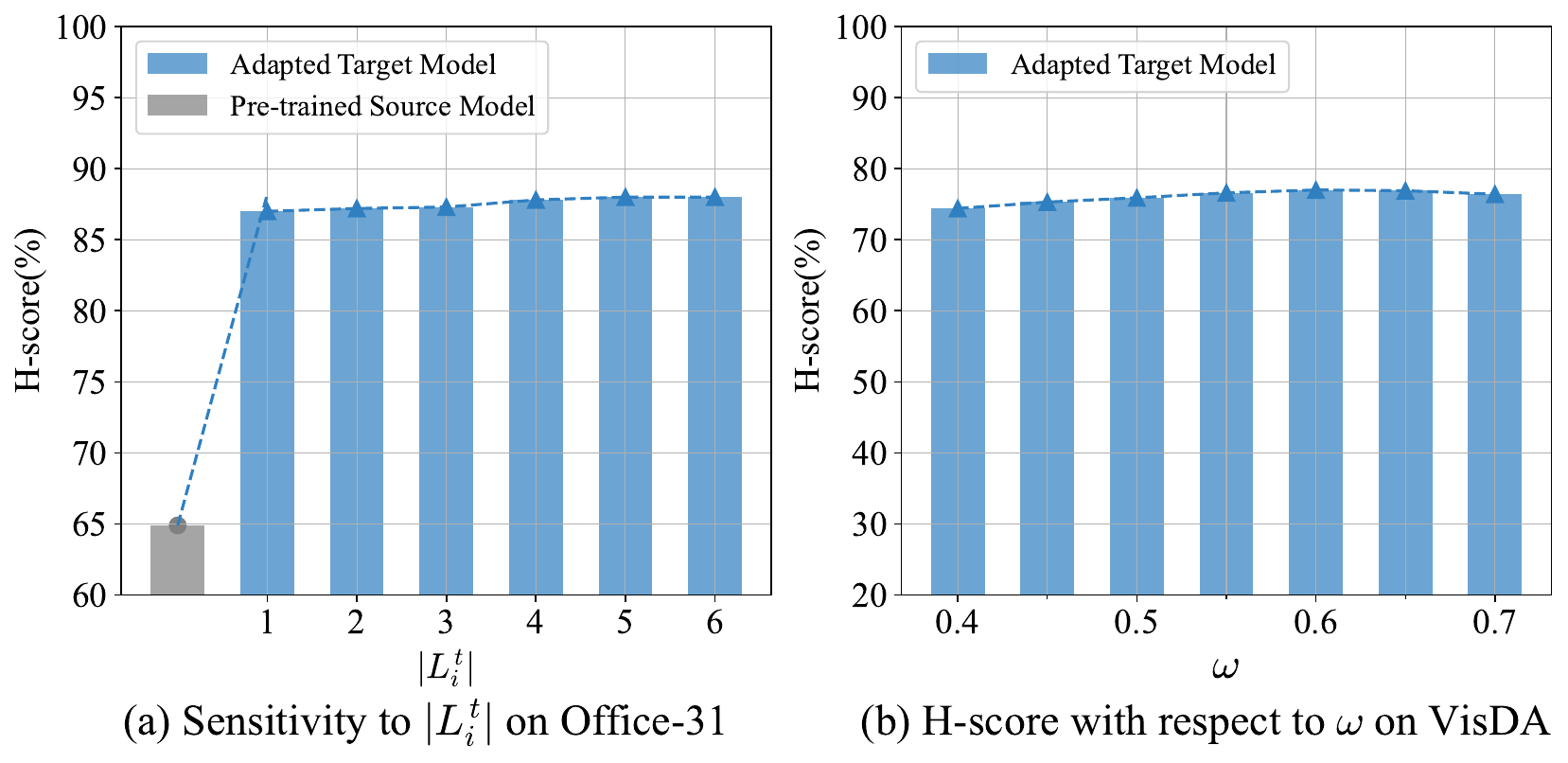}
    \vspace{-0.1in}
    \caption{\textbf{More Robustness Analysis}. (a) shows the sensitivity to $|L^t_i|$ on Office-31 in the OPDA scenario. (b) illustrates the H-score with respect to $\omega$ on VisDA in the OPDA scenario.}
    \vspace{-0.25in}
    \label{fig: analysis_3}
\end{figure}

\noindent \textbf{More Robustness Analysis.}  In addition to the robustness analysis presented in the main paper, we extend our investigation to the parameter sensitivity of the nearest neighbor set size $|L^t_i|$ in the context of the feature consensus learning objective $\mathcal{L}_{con}$. This analysis is conducted on the Office-31 dataset within the OPDA scenario, where the range of $\lambda$ is set to $[1, 2, 3, 4, 5, 6]$. As depicted in Figure~\ref{fig: analysis_3} (a), the results indicate that our LEAD framework demonstrates stability around the chosen parameter value of $|L^t_i| = 4$. Additionally, in Figure~\ref{fig: analysis_3} (b), we further analyze the H-score in relation to the threshold $\omega$ on the VisDA dataset, also in the OPDA scenario. It demonstrates a relative stability of the H-score around $\omega = 0.55$. Through an oracle validation, we could even achieve better performance when $\omega = 0.60$.

{
    \small
    \bibliographystyle{ieeenat_fullname}
    \bibliography{main}
}


\end{document}